\PassOptionsToPackage{final}{graphicx}
\PassOptionsToPackage{final}{listings}
\documentclass[a4paper]{llncs}

\usepackage[usenames,dvipsnames]{color} 

\usepackage{hyperref}       
\usepackage{url}            
\usepackage{amsmath,amssymb,amsfonts}       
\usepackage{booktabs}       
\usepackage{xcolor}
\usepackage{xspace}
\usepackage{todonotes}
\usepackage{framed}
\usepackage{paralist}
\usepackage[capitalize]{cleveref}
\usepackage{datetime2}
\usepackage{subcaption}
\usepackage{cases}
\usepackage{pgfplots}
\pgfplotsset{compat=newest}
\usepackage{wrapfig}
\usetikzlibrary{fit}

\DeclareRobustCommand{\cpp}
{\valign{\vfil\hbox{##}\vfil\cr
   \textsf{C\kern-.1em}\cr
   $\hbox{\fontsize{\sf@size}{0}\textbf{+\kern-0.05em+}}$\cr}%
}

\selectcolormodel{cmy}

\definecolor{lgrey}{rgb}{0.8,0.8,0.8}
\definecolor{grey}{rgb}{0.5,0.5,0.5}

\definecolor{lightblue}{rgb}{0.8,0.8,1.0}
\definecolor{lightred}{rgb}{1.0,0.8,0.8}
\definecolor{lightgreen}{rgb}{0.8,1.0,0.8}
\definecolor{angrygreen}{cmyk}{0.279,0,0.91,0.08}
\definecolor{lightred}{rgb}{1.0,0.8,0.8}
\definecolor{pink}{rgb}{1.0,0.1,1.0}

\title{Correct-by-Construction Runtime Enforcement in AI -- A Survey}

\author{
Bettina Könighofer\inst{\star 1}\and 
Roderick Bloem\inst{1} \and 
Rüdiger Ehlers\inst{2} \and
Christian Pek\inst{3}  
}

\institute{
Graz University of Technology, Institute IAIK, Graz, Austria
\and
Clausthal University of Technology, Clausthal-Zellerfeld, Germany
\and
KTH Royal Institute of Technology, Stockholm, Sweden
\\
\email{bettina.koenighofer@iaik.tugraz.at, roderick.bloem@iaik.tugraz.at}\\
\email{ruediger.ehlers@tu-clausthal.de, pek2@kth.se}
}

\begin{document}

\maketitle

\begin{abstract}
	Runtime enforcement refers to the theories, techniques, and tools for
enforcing correct behavior with respect to a formal specification of systems at runtime.
In this paper, we are interested in 
techniques for constructing runtime enforcers
for the concrete application domain of enforcing safety in AI.
We discuss how safety is traditionally handled in the field of AI and how more formal guarantees on the safety of a self-learning agent can be given by integrating a runtime enforcer. We survey a selection of work on such enforcers, where we distinguish between approaches for discrete and continuous action spaces.
The purpose of this paper is to foster a better understanding of advantages
and limitations of different enforcement techniques, 
focusing on the specific challenges that arise due to their application in AI. 
Finally, we present some open challenges and avenues for future work.

\end{abstract}

\begin{keywords}
 formal methods, runtime enforcement, shielding,  safety in AI, reinforcement learning
\end{keywords}

\setlength{\intextsep}{0pt}%
\section{Introduction}

Safety of learning-based or learned controllers is a major concern when using them in physical systems and in the proximity of humans.
Particularly during the exploration phase of learning, when an agent chooses random
actions to examine its surroundings, it is important to avoid actions that may
cause unsafe outcomes. 

Formal runtime enforcement techniques aim at guaranteeing safe
execution of learning-enabled systems. Due to their applicability to AI-based agents, they gain more and more attention. 
Such an enforcer not only detects unsafe choices by an agent, but also corrects them, so that the learning process can continue without interruption.

Which approaches are suitable for enforcing safety depends on the  environment in which the learner operates and the  challenges induced by the environment.
First, it is often the case that a learner interacts with a physical system, which induces the need to detect the unsafe behavior \emph{early enough to prevent it}. For dynamical systems, doing so involves planning ahead for a sufficient time span, whose length may not even be known. As a second challenge, many runtime enforcers must be \emph{real-time capable}. 
In particular, when analysing the safety of an agent's actions, a runtime enforcer cannot halt the 
environment, but rather must immediately decide whether the action issued by the agent should
be executed in the environment or replaced by a safe one.
A third challenge for runtime enforcement is that the dynamics of the \emph{environment may be partially unknown} before the learning process starts. This makes guarding the system's behavior against unsafe behavior difficult, as it is then unknown exactly which actions by an agent are unsafe. Thus, runtime enforcement may have to base on data-driven model identification.

In this paper, we give an overview of state-of-the-art correct-by-construction runtime
enforcement techniques that address the challenges that arise from the application of runtime enforcement to AI.

We start by discussing a diverse set of techniques developed in the field of AI addressing uncertainty in learning to increase the chance of the learning system to behave safely. 
These techniques tend to address the safety problem from a learning perspective aiming to increase the observed probability of safe system behavior. We discuss such approaches in Section~\ref{sec:secTwo}.

To keep systems safe with provable guarantees, more formal approaches are necessary. Since a learning system's behavior is not known in advance (and often randomized), the learner's safety cannot be verified upfront even if it were safe. Rather, the concept of \emph{runtime monitoring} is used to observe a learner's behavior, coupled with correcting the learner's choices whenever needed.
How to monitor systems is studied in the field of \emph{runtime verification} \cite{DBLP:series/lncs/10457}, which 
transfers concepts from the field of formal verification to testing a system's correctness at runtime. 
Classical runtime verification approaches can often deal with complex temporal specifications and are applied whenever a system is too complex to be analyzed upfront or when errors induced by the employed hardware are to be detected as well.

The majority of approaches from runtime verification are purely \emph{trace-based} and do not take the interaction between environment and system to be monitored into account to decide on when an error has been reached. This makes sense for most applications -- the possibility for the environment to force the system to violate its specification is often irrelevant as the environment itself only exhibits certain behaviors, for which the system may work correctly. 
This trace-based view is also present in the \emph{runtime enforcement literature} \cite{DBLP:journals/fmsd/PinisettyPTJFM17,DBLP:journals/fmsd/FalconeMFR11,DBLP:journals/tecs/PinisettyRSATH17} for non-AI-based systems, where the behavior of a system is only modified once the trace observed would start to violate the specification. When monitoring self-learning systems, this may however be too late to guarantee safety. 
This problem is addressed in some more recent works that employ a game-based perspective on monitoring \cite{DBLP:conf/rv/EhlersF11} and enforcement \cite{DBLP:journals/fac/RenardRF20}, which is suitable for settings with known environment capabilities.
Approaches for runtime enforcement of AI-based agents base on such ideas, can be broadly distinguished by whether they address \emph{discrete} or \emph{continuous state spaces}, and are addressed in Section~\ref{sec:secThree} and Section~\ref{sec:secFour} respectively.

In Section~\ref{sec:secThree}, we discuss approaches for \emph{discrete agent-environment systems}.
The state space of the environment and the action space of the agent are modelled to be discrete and finite, sometimes with  stochastic transitions.
In this context, runtime enforcement techniques can be further distinguished by whether unsafe actions are blocked by the learner or are used
for reward shaping. Differentiating aspects of approaches for runtime enforcement in discrete agent-environment systems are the different interference and communication techniques of the runtime enforcer with the agent, the used formal specification language, the used runtime enforcer synthesis techniques, and whether the safety analysis of actions is computed offline or online.
We discuss several recent works for discrete agent-environment systems
with a focus on advantages, limitations, and potential in their application to AI.

In Section~\ref{sec:secFour}, we discuss runtime enforcement techniques for \emph{continuous and 
hybrid agent-environment systems}. 
These systems pose additional challenges since they operate in continuous time, the state/action spaces are continuous, and the system dynamics are usually more complex, e.g., by having non-linearities or jumps between different modes of the hybrid system.
These challenges increase the effort to monitor such systems and to provide an accurate safety analysis (i.e., such that safety is guaranteed while the enforcer is comparably permissive towards the agent).
Such systems are even more susceptible to the curse of dimensionality (i.e., the state space of the system becomes unwieldy large) than discrete agent-environment systems. Hence determining proper abstractions and approximations of the system is even more important, particularly  if the safety analysis is to be performed in real time. 
For the surveyed approaches that base on the automatic computation of such abstractions or approximations, we discuss how they address this challenge.


After giving an overview of a good number approaches for runtime environment in AI-based systems, we then summarize some observations in Section~\ref{sec:secFive}, followed by the identification of future research directions.

\section{Safety in Data-Driven AI Methods}
\label{sec:secTwo}

Increasing the safety of AI approaches gains more and more attention within the AI community \cite{Schwalbe2020,Pereira2020}. 
Besides classical approaches, such as improving training data, removing biases, improving architectures and (hyper-)parameters, they aim to address safety from a system-level perspective. 
We can roughly categorize recent state-of-the-art approaches into three clusters:
1) understanding the capabilities and limitations of the AI system; 2) detecting and rejecting out-of-distribution/outlier inputs to the AI system; and 3) increasing the robustness of the AI system against unseen inputs and making decisions/actions of the system interpretable.
In the following paragraphs, we briefly review existing approaches in each of the categories.

\paragraph{Understanding the AI system:} The training data of an AI system consists of a set of examples (pairs of inputs and outputs) that solve the desired task.
Ideally, these examples are informative enough so that the system can generally solve the task, i.e., correctly mapping any input data to desired output data.
Yet, real-world training data may be far from being \emph{perfect}, resulting in approximation errors and uncertain mapping results of the system \cite{Papadopoulos2001,Guo2017,Toubeh2019}.
To estimate the residual uncertainty or prediction variance in the AI system, popular approaches use Bayesian methods \cite{Jensen1996,Kendall2017,Sensoy2018}, apply Monte Carlo dropout \cite{Globerson2006,Gal2016}, or utilize Deep Ensembles \cite{Lakshminarayanan2016,Guo2017}.
Yet, these uncertainty estimates are still not reliable, since perfect estimates can only be done given an infinite amount of data samples.

\paragraph{Outlier detection:} Uncertainties within an AI system can also be due to a distributional shift in which the data distributions are different in the training and operation domain \cite{Mohseni2019}.
For instance, the training dataset may lack corner cases that the system will encounter.
Moreover, learned features, such as the shape or color of objects, may look different in the operating domain of the AI system.
In general, these out-of-distribution (OOD) inputs may be detectable as outputs with a large prediction variance, but they can also be undetectable by having a low prediction variance.
Thus, it is important to already detect OOD inputs before feeding them to the system or explicitly consider them within the system.
The detection can be done, e.g., by including prediction-confidences in the network architecture \cite{Devries2018}, employing classifiers \cite{Vyas2018} and outlier detectors \cite{Hendrycks2018}, monitoring neuron activity outside the training data ranges \cite{Henzinger2019}, using self-supervised representation learning \cite{Golan2018}, active monitoring with human input \cite{Lukina2021}, incorporating temperature scaling \cite{Liang2017} or by incorporating generative adversarial networks \cite{Nitsch2020}.
OOD detection is very powerful, but often requires additional data or domain knowledge, rendering OOD detection infeasible for some applications. 

\paragraph{Robust and interpretable AI:} Yet another way to increase the safety of the AI system is to enhance its robustness and resiliency against unseen/OOD inputs, perturbations or corner cases \cite{Goodfellow2016}.
The overall aim is to improve the system's generalization capabilities. 
These improvements can be made through transfer learning \cite{Hendrycks2019}, regularization of the network \cite{Zhang2017}, verifying the networks robustness \cite{Lukina2021}, data augmentation \cite{Dreossi2018}, generative models \cite{Nalisnick2018,Mohamed2016,Ghadirzadeh2020}, or removing non-robust features from the training dataset \cite{Ilyas2019}.
Generalization improves robustness, yet there are no automated approaches to make a system more general or to check whether a system is general enough to not cause unsafe outputs. 
Interpretability approaches aim at making the output generation traceable \cite{Linardatos2021,Carvalho2019}.
For instance, by illustrating and explaining how the system derived a decision \cite{Selvaraju2016,Selvaraju2017,Mitsioni2021,Ribeiro2016,Amir2018}. 
As a result, (non-)experts are able to check whether the system can actually justify its decision.





Although the presented approaches increase the safety of AI systems and help to get a better understanding of the systems' limitations, these approaches cannot provide strict safety guarantees or even soft probabilistic guarantees \cite{Seshia2016}. 
Moreover, similar to testing-based validation, these approaches can often only reveal the presence of failures but not their absence. 
Formal runtime enforcement approaches aim specifically at providing strict safety guarantees by leveraging formal analysis techniques.


\section{Runtime Enforcement in Discrete Domains}
\label{sec:secThree}

In \emph{reinforcement learning} (RL)~\cite{sutton1998reinforcement}, an
agent aims to compute an optimal policy that maximizes the expected total amount of reward
received from the environment.
RL algorithms can be mainly divided into two categories: \emph{model-based RL} and \emph{model-free RL}.
In the model-based approach, 
the learning agent constructs a model of its environment in form of a Markov decision process (MDP).
By learning the model and applying planning approaches on the model, model-based RL can quickly obtain optimal policies, but the approach becomes impractical as the state space and action space grows. In contrast, model-free RL does not try to understand the environment and aims to learn a task through trial-and-error via interactions with the environment. Model-free RL is very scalable and has successfully been applied in solving various complex tasks, from playing video games to robotic tasks, but requires many samples for good performance.
Therefore, safety is an especially challenging problem for model-free RL, since the learning agent needs to explore many unsafe behaviour in order to learn that it is unsafe.
As a consequence, most formal runtime-enforcement techniques focus on ensuring safety in model-free RL, either during training or after training, or for both.

In most work, safety properties are formulated in linear temporal logic (LTL).
Several works consider only simple \emph{invariant safety properties} like  ``two robots should never collide''~\cite{GiacobbeHKW21}, other works consider the full safety fragment of LTL, 
which allows to formulate temporal safety properties like "whenever the first signal rises, the second signal has to rise within the next 5 time steps".

There are two popular directions of research for runtime enforcement for model-free RL: (1) using the safety property to compute a  maximally permissive enforcer, often called a shield, or (2) using the safety property for reward shaping.

\emph{Safety via Shielding.}
Shields have been used in model-free RL to enforce safe operation of an agent during training and after training (safety is guaranteed as long as the shield is used).
A shield~\cite{DBLP:conf/tacas/BloemKKW15} can automatically be computed from a given safety LTL specification and a model that captures all safety-relevant dynamics of the environment. 
The synthesis approach constructs a \emph{safety game} from the safety specification and the environmental model. The \emph{maximally permissive winning strategy} of the safety game allows all actions that will not cause a safety violation on the \emph{infinite horizon} and is implemented in the shield. 
Shields have been categorized in post-shields and pre-shield~\cite{DBLP:conf/aaai/AlshiekhBEKNT18}.
Post-shields monitor the actions selected by the agent and overwrite any unsafe action by a safe one.  Pre-shields are implemented before the agent and block, at every time step, unsafe actions from the agent (also referred to as action masking). At every time step, the agent can only choose from the set of safe actions.

Jansen et al.~\cite{DBLP:conf/concur/0001KJSB20} considered shielding in scenarios that incorporate uncertainty and therefore safety cannot be guaranteed. 
They introduce the concept of a probabilistic
shield that enables RL decision-making to adhere to safety constraints with high probability.
Probabilistic model checking techniques are used to compute the probabilities of all states and actions of the MDP to satisfy a safety LTL property, called the safety value of an action.
A shield blocks an action if its safety value is smaller than some absolute threshold or relative threshold.
Considering safety as quantitative measure allows risk taking and to tune the trade-off between safety and performance. 
Giacobbe et al.~\cite{GiacobbeHKW21} applied the same technique on Atari 2600 games and specified 43
safety properties for 31 games. The authors computed shields using a bounded horizon for all properties. Applying these shields resulted in the safest RL agents for Atari games
currently available.

ElSayed-Aly et al.~\cite{DBLP:conf/atal/Elsayed-AlyBAET21} considered shielding for multi-agent reinforcement learning (MARL) and proposed
to either synthesize a single centralized shield that monitors and corrects all agents’ joint actions,
or to synthesize multiple shields where each shield is only responsible for a subset of agents at each step.
Furthermore, the authors introduced a minimal interference criteria for the MARL setting: 
a shield should change the actions of as few agents as possible when correcting an unsafe joint action.

To compute a shield upfront, the safety of all actions for all reachable states has to be analyzed and stored.
For large environments, this results in long offline computation times and huge shielding data bases,
rendering shielding not tractable, and for dynamic or partially unknown environment, the offline computation is not possible.
To tackle this issue, Könighofer et al.~\cite{DBLP:conf/nfm/KonighoferRPTB21} perform the safety analysis of the actions \emph{online}.
Using the time between two successive decisions of an agent, their approach builds an MDP quotient that captures the behaviour of the environment in the next $n$ steps and analyzes the actions for the next decision states on the fly.
However, the approach does not provide any worst-case guarantee on the computation time of the safety analysis. Therefore, the approach is suited for settings in which the agent does not have to make a decision in every time step and if needed, halting the agent to wait for a decision of the shield does not cause any harm.

Achiam et al.~\cite{DBLP:conf/icml/AchiamHTA17} encoded safety in terms of constraints, leading to the line of research on \emph{constrained policy optimization}. Constrained Markov decision processes (CMDPs) are used to decouple safety from reward, where an independent signal models the safety aspects. An optimal policy balances the trade-off between safety and performance, but there are no safety guarantees during learning. 
Therefore, this line of work was extended by Simão et al.~\cite{DBLP:conf/atal/SimaoJS21}
by using a factored MDP that represents only the safety aspects of the full CMDP. 
The factored MDP is used to restrict the exploration of the learner and thereby 
allowing the RL agent to learn an optimal policy for the CMDP without violating the
constraints.

\emph{Safety via Reward Shaping.}
Several recent works~\cite{HahnPSS0W20,DBLP:journals/corr/abs-1902-00778} 
use safety properties expressed in LTL for reward shaping and are often referred to as \emph{logically-constrained RL}. Approaches based on reward shaping will result in agent that will minimize the risk of safety violations after training.
To ensure safety during training, such approaches need to be extended by restricting the exploration during training~\cite{HasanbeigAK20}. 

In logically-constrained RL, a formula expressing some desired properties is first converted to an automaton, which is then translated into a state-adaptive reward structure. Any RL-agent trained with this reward structure results in policies that maximise the probability of satisfying the given formula. In the context of safe RL, the formula expresses a safety property and the trained agent will minimize the risk of violating the property. 

Most works mentioned in this section discuss the impact of enforcing safety on the learning performance
and provide requirements that the enforcement mechanism needs to satisfy to preserve the 
convergence guarantees of the learner~\cite{DBLP:conf/atal/SimaoJS21,DBLP:conf/aaai/AlshiekhBEKNT18}. Several papers show empirically that enforcing safety during learning has the potential to increase the agent's performance if the safety and the performance properties are aligned.

\section{Runtime Enforcement in Hybrid/Continuous Domains}
\label{sec:secFour}

Ensuring the correct behavior of a system is particularly important for \emph{safety-critical} systems. These operate in the physical world, which means that quantities of space and time become continuous. This complicates the enforcement of properties of such systems -- as many verification problems for hybrid systems are undecidable, maximally permissive enforcement is undecidable in the general case, too. At the same time, many optimization problems for system behavior are equally undecidable in such environments and the aspects of the environment dynamics that are relevant for optimizing the system behavior can be unknown a-priori, which makes reinforcement learning attractive to allow an agent to adapt to the environment dynamics.

Multiple streams of research have emerged that circumvent the undecidabilty of maximally permissive enforcement in various ways, such as focusing on simpler system dynamics and restricting the system's behavior in ways that are not maximally permissive.

A good example of the first stream of research is the work by Goorden et al.~\cite{DBLP:conf/adhs/GoordenLNNRS21}, in which a maximally permissive controller is computed from a model in which the discrete behavior of the system may only depend on clock variables, so that controller synthesis algorithms from the area of timed automata can be used. This maximally permissive controller is then used as a system behavior constraint in a reinforcement learning process. 

Another example is a safe learning approach by Perkins and Barto \cite{DBLP:journals/jmlr/PerkinsB02},
in which they propose restricting a safe learning process to letting an agent learn how to switch between different Lyapunov-stable control laws that are known to be safe even when switching between them in an arbitrary manner. 

To handle also complex learner's behavior in complex environments, the learners can be restricted to safe behavior in a way that is not maximally permissive. 
For instance, Fisac et al. \cite{Fisac2018} present an approach to compute controlled invariant sets for known system dynamics that is based on a closed-form characterization of this set, which encodes the least-restrictive control law, which is subsequently approximated numerically (in the case of complex system dynamics). 

Another approach is to synthesize control barrier functions for known system dynamics, as done by Cheng et al.~\cite{DBLP:conf/aaai/ChengOMB19}. These lead to guaranteed safety of the learner when restricting it so that the learner never uses an action not allowed by the control barrier. 
The drawback of their approach is that the computed control barrier functions have to be affine, so that in order for them to be safe, they have to be computed in a way that makes them overly conservative. On the plus side, affine control barrier functions are easy to compute and completely continuous.

Fulton and Platzer \cite{DBLP:conf/aaai/FultonP18} provide an approach called \emph{Justiﬁed Speculative Control} that operates by using a provably correct runtime monitor to filter the unsafe actions of a system and enforce that the learner only takes safe actions. Their approach builds on a provably correct non-deterministic backup strategy, for which the learner has to select actions allowed by a  boundary strategy until it (possibly) becomes apparent that the environment of the learner does not behave in the way assumed for the formal safety proof of the boundary strategy (in which case the learner can also potentially choose unsafe actions). Their approach has the interesting property that it can also be applied if the backup strategy is only defined and verified for a part of the environment's state. In this case, by integrating a distance metric to the modeled states into the learning process, the learner can be guided back to the safe states, which reduces the probability of unsafe behavior even for partially unknown environments.

Nageshrao et al.~\cite{DBLP:conf/smc/NageshraoTF19} discuss an approach to integrate so-called ``short horizon safety checks'' into a reinforcement learning approach. Learner actions that are unsafe according to these checks are altered. They also record whenever this happens so that the unsafe situation can be replayed to the learner to speed up the convergence of the learner's policy to a safe one.

Since self-learning systems are particularly interesting for unknown system dynamics, there is also a rich body of works dealing with the question of how to enforce the safety of a learner in this case. The starting point is usually that the environment dynamics are only \emph{partially known}, as if they are fully unknown, it is not even possible to guarantee the safety of the learner with the first step of the system's execution.

For instance, Gillulga and Tomlin \cite{DBLP:conf/rss/GillulaT12} define an approach in which a self-learning controller is embedded into a safety controller that enforces the safety of a controlled physical system. In the approach, reachability analysis for the physical system is used in tandem with observing the disturbances observed at runtime, so that the safety wrapper becomes neither too conservative nor too permissive.

Then, the safety enforcement approach by Fisac et al. \cite{Fisac2018} mentioned above also includes a means to identify the magnitude of disturbances at runtime to adapt the enforcer at runtime to avoid being unnecessarily conversative. Of course, this means that if disturbances perform a sudden increase, this leads to temporarily unsafe behavior. Learning the unknown aspects of the environment dynamics is also present in other work.

Cheng et al. \cite{DBLP:conf/aaai/ChengOMB19} provide an approach for safe reinforcement learning that bases on the use of control barrier functions to detect unsafe actions by a learner as well as to guide the behavior of the learner.
Such control barrier functions can be used under uncertain system dynamics, but the authors also include a process to learn a more precise model of the system dynamics over time, which enables a refinement of the control barrier functions to become less conservative when more information on the environment dynamics becomes available.

\section{Discussion on Runtime Enforcement in AI}\label{sec:conclusion}
\label{sec:secFive}

\subsection{Observations from the State of the Art}

In a variety of applications, AI systems will only unfold their full potential if such systems are safe during their operation.
In our overview, we observed that the formal methods and AI communities often address safety from different perspectives. 
Formal runtime verification approaches aim at providing formal or probabilistic guarantees at all times, whereas AI-based safety approaches focus on improving safety as much as possible while not decreasing the system's performance, e.g., by detecting out-of-distribution data or enhancing the system's overall robustness.
Formal methods usually regard the AI system as a black box whose 
actions might be adversarial. 
To detect (potentially) unsafe 
actions while lowering computational efforts, runtime verification approaches usually make use of abstractions or simplified models of the AI system (as, e.g., in \cite{DBLP:conf/aaai/AlshiekhBEKNT18}).
However, such simplifications might result in more conservative behavior of the system \cite{Brunke2021} or be even incorrect for complex applications.
AI-based approaches, on the other hand, commonly strive for model-free solutions so that the system is as free as possible to learn the desired task.

Specifically for formal methods approaches, we saw that significant advances have been made in recent years, from introducing new theory to applying runtime verification to highly complex AI systems.
Yet, we also observed that the conducted experiments are hard to compare with each other. 
In contrast to machine learning, the formal methods community has only first traces of a standard benchmark set \cite{fmrChallenge} that allows one to compare the results of published approaches with each other. This lack of a big standardized benchmark set possibly hinders the community to advance theory but also to provide significant practical contributions.

This problem is caused by different problem domains, various challenges within each application, as well as differing technical foundations in the runtime verification approaches. For instance, techniques based on reachability checking are hard to compare against barrier function-based approaches.

Moreover, the common metrics to assess the performance of runtime verification approaches (e.g., runtime overhead in runtime monitoring or the expressivity of the specification language, \cite{DBLP:journals/fmsd/SanchezSABBCFFK19}) are not appropriate when an unavoidable degree of conservativeness is used in monitoring  (black-box) AI systems. 
Hence, in order to establish a common comparison, new metrics are needed.
However, we see that more and more research groups are investigating these challenges and contributing towards enforcing safety in AI-based systems.


\subsection{Future Directions and Conclusions} 

Our communities need to overcome several challenges to make AI systems safe to use. 
Most AI approaches aim for being free of any model; yet, most formal methods approaches require a model.
Model identification for large systems is still a challenging task and for certain applications, e.g., a robot that needs to make contact with a human, it is even difficult to design neural networks that approximate the dynamics or to mathematically define safety \cite{Mitsioni2019,Zhang2017b}. 
We need to explore new techniques to analyze AI systems and to automatically generate models or abstractions of them.
Moreover, many AI systems will be deployed in uncertain and partially observable environments, such as in autonomous driving.
In such environments, it is even more challenging to verify safety, since the environment might be unknown and needs to be explored online, the behavior of agents may be unknown, and information on the system may be partial and even incorrect. 
We need more research in analyzing and exploring the environment and system on-the-fly during its operation. 
To make our research more comparable, our communities need to come up with challenging benchmarks that help us to measure the performance of our algorithms as well as to advance theory. 
In robotics and AI, the availability of benchmarks have led to new research that  drastically improves the performance of algorithms \cite{Moll2015,Sucan2012,Duan2016}. 
Even though environments such as the OpenAI safety gym  \cite{Ray2019} exist, they are made from the perspective of the AI community and do not reflect the requirements for runtime enforcement.
Finally, both the AI and formal methods communities need to closely work together and foster synergies.
In this way, we can increase safety, performance, and interpretability of AI-based systems together. 

    

%
%


\bibliographystyle{abbrv}
\bibliography{literature}

\end{document}